\RequirePackage{fix-cm}
\documentclass{article}
\usepackage[preprint]{neurips_2026}

\usepackage[utf8]{inputenc} 
\usepackage[T1]{fontenc}    
\usepackage{hyperref}       
\usepackage{url}            
\usepackage{booktabs}       
\usepackage{amsfonts}       
\usepackage{nicefrac}       
\usepackage[export]{adjustbox}
\usepackage{tcolorbox}
\usepackage{microtype}      
\usepackage{xcolor}         
\usepackage{multirow}
\usepackage{graphicx}
\usepackage{adjustbox}
\usepackage{amsmath}
\usepackage{wrapfig}
\usepackage{tabularx}
\usepackage{array}

\usepackage[table]{xcolor}
\usepackage{booktabs}
\usepackage{multirow}
\usepackage{subcaption}

\definecolor{hA}{HTML}{63BE7B} 
\definecolor{hB}{HTML}{B1D580} 
\definecolor{hC}{HTML}{FFEB84} 
\definecolor{hD}{HTML}{FCB47A} 
\definecolor{hE}{HTML}{F8927B} 
\definecolor{hF}{HTML}{F8696B} 

\newcommand{\hcell}[2]{%
  \ifdim #1pt>0.95pt \cellcolor{hA}#2\else
  \ifdim #1pt>0.85pt \cellcolor{hB}#2\else
  \ifdim #1pt>0.70pt \cellcolor{hC}#2\else
  \ifdim #1pt>0.55pt \cellcolor{hD}#2\else
  \ifdim #1pt>0.4pt \cellcolor{hE}#2\else
  \cellcolor{hF}#2\fi\fi\fi\fi\fi}

\newcommand{\h}[1]{\hcell{#1}{#1}} 
\newcommand{\hb}[1]{\hcell{#1}{\textbf{#1}}} 
\newcommand{\negd}[1]{\textcolor{red!75!black}{\textbf{#1}}}
\newcommand{\posd}[1]{\textcolor{green!50!black}{\textbf{#1}}}

\title{Dense Contexts Are Hard Contexts:\\Lexical Density Limits Effective Context in LLMs}

\author{%
  Giovanni Dettori \\
  Department of Computer Science\\
  Politecnico di Torino\\
  Torino, IT 10129 \\
  \texttt{giovanni.dettori@polito.it} \\
  \And
  Matteo Boffa \\
  Department of Computer Science\\
  Politecnico di Torino\\
  Torino, IT 10129 \\
  \texttt{matteo.boffa@polito.it} \\
  \And
  Danilo Giordano \\
  Department of Computer Science\\
  Politecnico di Torino\\
  Torino, IT 10129 \\
  \texttt{danilo.giordano@polito.it} \\
  \And
  Idilio Drago \\
  Department of Computer Science\\
  University of Turin\\
  Torino, IT 10124 \\
  \texttt{idilio.drago@unito.it} \\
  \And
  Marco Mellia \\
  Department of Computer Science\\
  Politecnico di Torino\\
  Torino, IT 10129 \\
  \texttt{marco.mellia@polito.it} \\
}
\begin{document}
\maketitle
\begin{abstract}
  
Input length and the position of relevant information are widely cited as the primary causes of degraded LLM long-context performance. 
Here, we study \emph{lexical density} -- the rate at which a context introduces distinct information -- as a third, largely overlooked factor that systematically reduces the effective context window of LLMs.
We quantify the impact of lexical density on open-weight LLMs (9B–685B) using three ``find-the-needle'' style benchmarks with identical length ($\approx$12k tokens) and controlled needle position, but increasing density of information. We observe a sharp performance collapse in higher-density benchmarks: models that are near-perfect in sparse contexts drop below 60\% retrieval score on denser ones. 
To rule out task-type confounds, we vary and control the density within each benchmark while keeping all other properties unchanged. Reducing density generally restores performance, especially in the high-density regimes where degradation appears. 
These results show that effective context capacity is a function of lexical density, with direct implications for real-world LLM systems operating on compact, information-rich inputs.\footnote{Link to the code: \url{https://anonymous.4open.science/r/LLM_Density-1AB7/}}

\end{abstract}
\section{Introduction}\label{sec:intro}

Modern LLM applications assemble inputs from increasingly heterogeneous sources -- user prompts, agent configurations, persistent memory, tool outputs, and retrieved documents~\citep{yao2022react,packer2023memgpt,lewis2020retrieval}.
However, reliably \emph{using} this large context remains an open problem: models often fail to incorporate task-critical evidence from their inputs, producing errors that propagate through downstream actions and are difficult to attribute. Hence, understanding \emph{when} and \emph{why} models fail to use their context is a prerequisite to build reliable LLM-based systems.

Two factors are primarily known to drive context-related failures: (i) \emph{Length}: as input grows, models become unreliable in locating relevant information, with performance often collapsing well before the advertised window~\citep{kamradt2023niah,hsieh2024ruler}; and (ii) \emph{Position}: information in the middle of the context is recovered less reliably than at the edges (the \emph{Lost in the Middle} effect)~\citep{liu2024lost}. Both effects intensify under distractors and heavier reasoning loads~\citep{modarressi2025nolima,chroma2025contextrot}. 
However, existing evaluations implicitly assume that contexts of equal length and layout are equivalent. In this paper, we show that this is not the case, as two contexts with identical length and the same relevant information placement can show sharply different levels of difficulty. We identify a third axis explaining this discrepancy: \emph{lexical density} -- the rate at which a context introduces distinct information.
Intuitively, redundant text can be skimmed, whereas dense text requires processing nearly every token.
 
We hypothesize that density reduces retrieval reliability at fixed length and position, shrinking the usable context window of LLMs. We quantify density using the Moving-Average Type-Token Ratio (MATTR)~\citep{covington2010mattr},\footnote{For reference, standard prose such as \emph{Pride and Prejudice} yields MATTR$\approx$0.72~\citep{austen1813pride_gutenberg}, while agentic configuration files such as OpenClaw Markdown Files can reach MATTR$\approx$0.82~\citep{berman2026openclaw_markdown_files}.} and test our hypothesis on \emph{eight open-weight LLMs (9B--685B parameters)} across three ``find-the-needle'' benchmarks. The benchmarks have identical length ($\sim$12k tokens), structure, and needle placement strategy, but varying density: MK-NIAH (MATTR=0.58)~\citep{kamradt2023niah}, and our two novel benchmarks \textsc{Scene-Rules} (MATTR=0.75) and \textsc{WordChecker} (MATTR=1.0, where nearly every token is new), introduced to explicitly control density.

\begin{figure}[t]
    \centering
    \includegraphics[width=\linewidth]{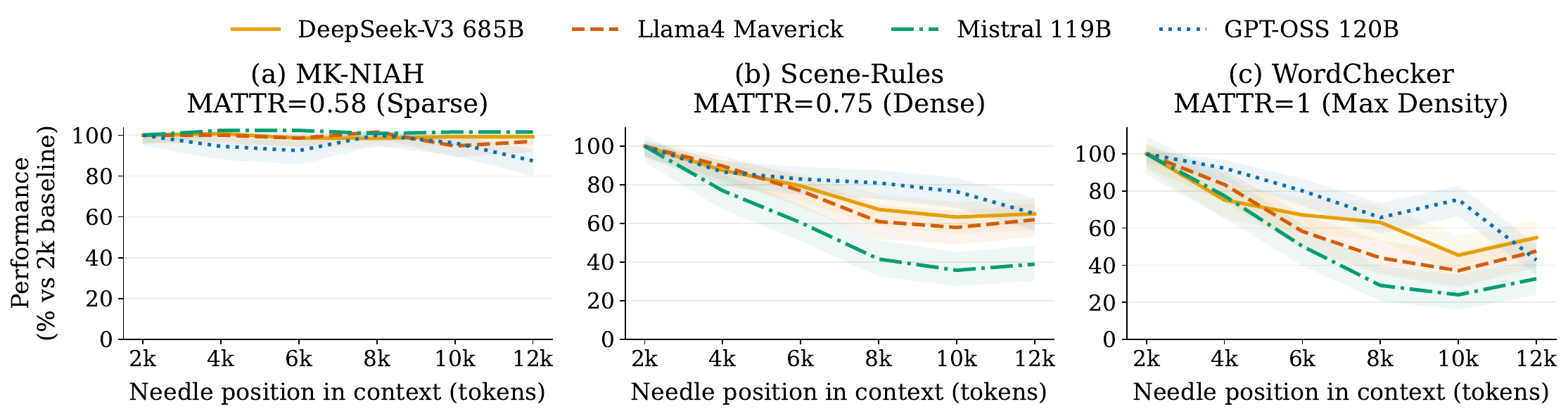}
    \vspace{-1.2em}
    \caption{\textbf{Retrieval score across needle positions, normalized to the scores at the 2k position.} Shaded bands indicate 95\% Wilson score intervals. All benchmarks use $\approx$12k-token contexts and matched needle placement, but differ in lexical density and retrieval form. The score remains high on sparse MK-NIAH, but collapses on denser Scene-Rules and WordChecker. \textbf{This motivates density as a complementary candidate axis of long-context degradation.}}
    \label{fig:fig_1}
    \vspace{-1.5em}
\end{figure}

Across benchmarks matched for length and needle placement, performance drops sharply in dense settings: models near-perfect on MK-NIAH collapse past 4k--6k tokens on Scene-Rules and WordChecker (Figure~\ref{fig:fig_1}). This represents a failure regime an order of magnitude earlier than predicted by length alone~\cite{hsieh2024ruler}. Since the three benchmarks solve different tasks -- exact-string, semantic, and lemma-aware -- we further isolate density with within-benchmark interventions: we artificially control density while we keep task, length, and position fixed. Reducing lexical density consistently restores performance, establishing density as an additional axis of long-context degradation alongside length and position.


\noindent\textbf{Contributions:}\quad (i) We identify lexical density as a previously unexplored axis in long-context evaluation and show that it systematically reduces effective context capacity. (ii) We introduce two new benchmarks, \textsc{Scene-Rules} and \textsc{WordChecker}, enabling controlled studies of density at fixed length and position. (iii) Across eight LLMs, we demonstrate that density induces performance degradation significantly earlier than predicted by length alone. 
(iv) Through within-benchmark interventions that hold the length, position, and task structure fixed, we show that varying lexical density produces large performance swings, ruling out length and position as sole explanations.

\section{Literature Review}\label{sec:lit_review}

\noindent\textbf{Long-context evaluation.}\quad
The Needle-in-a-Haystack (NIAH) paradigm~\citep{kamradt2023niah} has anchored long-context evaluation, with extensions varying task complexity~\citep{hsieh2024ruler}, pushing context to 128k--256k tokens~\citep{zhang2024inftybench,yuan2024lveval}, adding document-grounded, multi-fact, or interference settings~\citep{bai2024longbench,shaham2023zeroscrolls,kuratov2024babilong,xu2024taskhaystack}, and generalizing across modalities and languages~\citep{wang2024mmniah,kim2025oneruler}. A growing consensus holds that NIAH is \emph{saturated}: frontier models solve it near-perfectly regardless of length or position. The dominant response has been to change the \emph{task} -- toward application-centric~\citep{yen2025helmet}, retrieval-and-RAG~\citep{lee2025loft}, or latent-structure~\citep{vodrahalli2024michelangelo} settings. We argue that saturation also stems from an overlooked property of the setup itself: NIAH haystacks are typically lexically redundant. Increasing the haystack density, while holding the retrieval task fixed, suffices to reproduce many of the failures that motivated these task-level benchmarks.

\begin{wraptable}{r}{0.53\linewidth}
\vspace{-1em}
\centering
\small
\setlength{\tabcolsep}{4pt}
\renewcommand{\arraystretch}{1.05}
\caption{Axes of long-context degradation in prior work. Input \textbf{Len}gth. Needle \textbf{Pos}ition. \textbf{Q}uery--\textbf{N}eedle similarity. \textbf{Dist}ractor similarity. Haystack lexical \textbf{Den}sity, which we introduce.}
\label{tab:positioning}
\begin{tabular}{lccccc}
\toprule
 & Len & Pos & Q--N & Dist & Dens \\
\midrule
\citet{kamradt2023niah}      & \checkmark & \checkmark &            &            &            \\
\citet{hsieh2024ruler}       & \checkmark & \checkmark &            &            &            \\
\citet{liu2024lost}          &            & \checkmark &            &            &            \\
\citet{levy2024same}         & \checkmark &            &            &            &            \\
\citet{modarressi2025nolima} &            &            & \checkmark &            &            \\
\citet{chroma2025contextrot} &            &            &            & \checkmark &            \\
\textbf{Ours}                &            &  \checkmark          &            &            & \textcolor{red}{\checkmark} \\
\bottomrule
\end{tabular}
\vspace{-2em}
\end{wraptable}

\noindent\textbf{Decomposing long-context degradation.}\quad
Table~\ref{tab:positioning} summarizes the axes isolated by prior work; none captures haystack density. Position effects~\citep{liu2024lost,hsieh2024found} concern \emph{where} the needle appears, not the information load around it; we confirm needle-position effects and show they are amplified by density, emerging earlier in dense settings. \citet{levy2024same} isolate length by padding a fixed reasoning sample; we make the analogous move for density, holding length and position fixed and varying only the haystack. Most closely, \citet{modarressi2025nolima} and \citet{chroma2025contextrot} show that reducing query--needle or query--distractor lexical overlap degrades performance. Density generalizes this from a pairwise to a haystack-level property: when \emph{all} context items are mutually dissimilar, retrieval is hardest even at fixed query--needle similarity. The two effects are complementary.

\noindent\textbf{Information density as a prompt property.}\quad
That natural language carries non-uniform information per token is classical: Shannon noted the redundancy of English~\citep{shannon1951prediction}, and the Uniform Information Density literature posits that speakers modulate surface forms to keep per-token information roughly constant~\citep{levy2008expectation,jaeger2010redundancy}. Prompt-compression methods operationalize this by scoring and dropping low-information tokens or chunks~\citep{jiang2023llmlingua,jiang2024longllmlingua,li2023selective,pan2024llmlingua,chen2025dast}, replacing prompts with learned summaries~\citep{mu2023gist,chevalier2023adapting}, or framing compression as rate--distortion~\citep{nagle2024fundamental}. This literature treats density as a knob to \emph{turn down}. Our question is the inverse: when density is already high -- as in agentic configurations, structured memory, or tool outputs -- can the model still use the context?

\noindent\textbf{Positioning.}\quad
We model text-intrinsic lexical density via the Moving-Average Type--Token Ratio (MATTR)~\citep{covington2010mattr}, formalized in Section~\ref{sec:density}. Low MATTR indicates a redundant haystack; high MATTR, one where nearly every word is new. Computed from the haystack alone -- independent of query, needle, and relational structure -- MATTR is complementary to position~\citep{liu2024lost}, task complexity~\citep{hsieh2024ruler}, query--evidence overlap~\citep{modarressi2025nolima}, and distractor similarity~\citep{chroma2025contextrot}.
We empirically show that lexically dense haystacks induce retrieval failures past 4k tokens where sparse haystacks remain reliable: effective context capacity depends not only on token budget, but also on lexical load.
\section{Methodology}\label{sec:methodology}
\subsection{Benchmarks: MK-NIAH, Scene-Rules, and WordChecker}\label{sec:benchmarks}

\begin{table}[thb]
\vspace{-1.2em}
\centering
\scriptsize
\renewcommand{\arraystretch}{1.35}
\setlength{\tabcolsep}{2.5pt}

\caption{\textbf{Benchmark overview.} All benchmarks are cast as exact-retrieval tasks under an approximately 12k-token context budget. The table reports the query, candidate format, average candidate length \(\bar{|c|}\), number of candidates \(|C|\), and matching mechanism.}
\vspace{5pt}

\resizebox{\columnwidth}{!}{%
\begin{tabular}{@{}
    >{\raggedright\arraybackslash}m{0.13\columnwidth}
    >{\raggedright\arraybackslash}m{0.20\columnwidth}
    >{\centering\arraybackslash}m{0.24\columnwidth}
    >{\centering\arraybackslash}m{0.07\columnwidth}
    >{\centering\arraybackslash}m{0.07\columnwidth}
    >{\raggedright\arraybackslash}m{0.20\columnwidth}
@{}}
\toprule
\textbf{Benchmark} &
\textbf{Query} &
\textbf{Candidate Format} &
\textbf{$\bar{|c|}$} &
\textbf{\textbf{$|\mathcal{C}|$}} &
\textbf{Matching Mechanism} \\
\midrule

\textbf{MK-NIAH} &
Retrieve value for $u_j$ &
\includegraphics[width=\linewidth, valign=m]{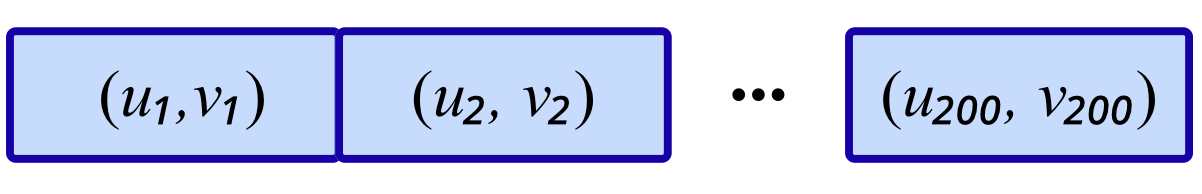} &
58 &
200 &
Exact-string (UUID) \\

\textbf{Scene-Rules} &
Find $r_i$ violating $s_i$ &
\includegraphics[width=\linewidth, valign=m]{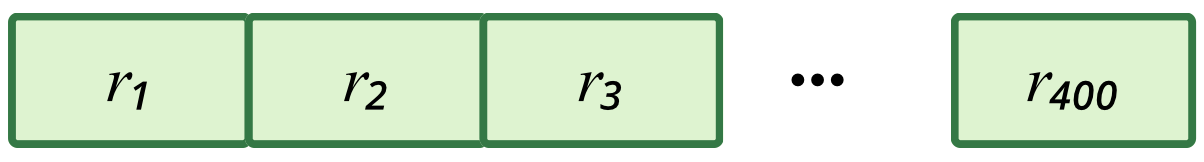} &
28 &
400 &
Semantic \\

\textbf{WordChecker} &
Find forbidden $w_j$ in $p_i$ &
\includegraphics[width=\linewidth, valign=m]{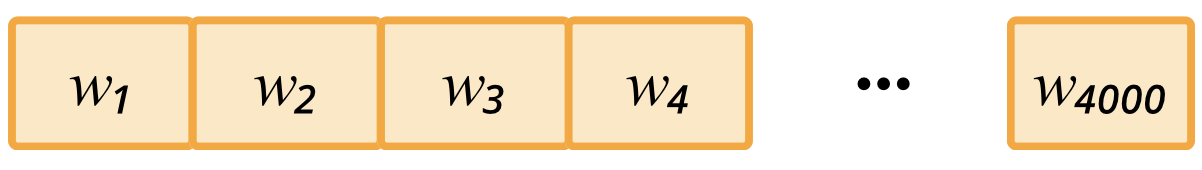} &
3 &
4000 &
Lemma-aware string \\

\bottomrule
\end{tabular}%
}
\label{tab:benchmark_taxonomy}
\end{table}

We design \emph{retrieval tasks} to study how lexical density affects an LLM's use of context. 
We cast all benchmarks in a common format. Each instance consists of a query \(q\), a candidate set \(\mathcal{C}=\{c_j\}_{j=1}^{m}\), and a unique target answer \(y^\star\), the \emph{needle}. The model receives a prompt 
\(x=\operatorname{prompt}(q,\mathcal{C})\) and outputs an answer \(\hat{y}=f_\theta(x)\), where $f_\theta(x)$ denotes the LLM’s generation function. 
The objective is exact retrieval: the prediction is correct if $\hat{y}$ matches the target 
$y^\star$.\footnote{We evaluate exact match (string match) between the $\hat{y}$ and $y^\star$. For WordChecker, both predictions and targets are normalised to their base lemmas prior to comparison.} We denote by 
\(|\mathcal{C}|\) the number of candidates and by \(|x|\) the length of the prompt in tokens. As candidates occupy most of the context, we approximate the prompt length as 
\(|x| \approx \bar{|c|} \cdot |\mathcal{C}|\), where \(\bar{|c|}\) is the average candidate length in tokens. 
Across all benchmarks, we fix context length ($|x|\approx12$k tokens), candidate structure, and retrieval format, while varying lexical density and needle position.
Unlike prior work using up to 256k tokens~\citep{zhang2024inftybench,yuan2024lveval}, we operate at $\approx$12k tokens, a regime not expected to induce degradation from length alone~\citep{hsieh2024ruler}. We describe the three benchmarks below, summarise details in Table~\ref{tab:benchmark_taxonomy}, and report examples in  Appendix~\ref{app:examples_benchmarks}. 

\noindent\textbf{Multi-Key Needle-in-a-Haystack (MK-NIAH).}\quad
MK-NIAH is the hardest variant of the standard Needle-in-a-Haystack task
\citep{hsieh2024ruler,yen2025helmet}. The candidate set consists of user--value records,
\(\mathcal{C}=\{(u_j,v_j)\}_{j=1}^{m},\)
where both \(u_j\) and \(v_j\) are UUIDs (35 character-long alphanumeric strings). The query \(q\) specifies one user ID. The model must locate the record whose user field matches the query and return the corresponding value. Specifically, there is a 
unique index \(j^\star\) such that \(u_{j^\star}\) matches the queried ID. The target answer is the associated value \(y^\star = v_{j^\star}\). The usage of UUIDs prevents semantic shortcuts and requires retrieval from the provided context. We saturate the approximately 12k-token budget setting \(|\mathcal{C}|=200\), as each user–value averages 58 tokens.

\noindent\textbf{Scene-Rules.}\quad
Scene-Rules is the first new benchmark we introduce. Starting from the Moral Stories dataset
\citep{emelin2020moralstoriessituatedreasoning}, we construct scenario-rule ($S-R$) pairs where the scenario violates exactly one rule (see Appendix~\ref{app:examples_benchmarks}). We turn this into a retrieval task by setting the query to the scene, \(q=s_i\), and the candidate set to a subset of rules, \(\mathcal{C}=\mathcal{R}_i \subset R,\) where \(r_i \in \mathcal{R}_i\) is the unique gold candidate. The model must identify which rule in \(\mathcal{R}_i\) the scene violates, and retrieve the rule (\(y^\star = r_i\)). 
Unlike MK-NIAH, this task requires semantic matching between the scene and the candidate rules. Each rule accounts for 28 tokens on average; therefore, we set \(|\mathcal{C}|=400\) to saturate the 12k-token budget.

\noindent\textbf{WordChecker.}\quad
WordChecker is the second new benchmark we introduce, inspired by
\citet{lin-etal-2020-commongen,boffa2025large}. The query \(q\) is a phrase $p_i$, and the candidate set is a list of forbidden words, \(\mathcal{C}=W_i=\{w_j\}_{j=1}^{m}.\)
Exactly one candidate word  \(w^\star \in W_i\) appears in the phrase $p_i$, under lemma normalisation. The model must retrieve such a forbidden word from the candidate list (\(y^\star = w^\star\)). Compared with the previous benchmarks, the candidates are much more granular: words account for 3 tokens on average. Hence, we set \(|\mathcal{C}|=4000\) to fill the 12k-token budget. This task does not require high-level semantic reasoning; it is a lemma-aware string-matching problem. 

The three benchmarks differ along two axes: lexical density (our object of study) and the type of query–candidate matching required (exact, semantic, and lemma-aware). In Section~\ref{sec:manipulation}, we isolate the effect of density through a within-benchmark intervention that varies density while holding the matching task, context length, and needle position fixed. Because such manipulations can introduce artefacts (Section~\ref{sec:decoupling}), we design benchmarks with uniform candidate structure to better control them -- an assumption not satisfied by existing long-context tasks. We discuss residual artefacts in Section \ref{sec:limitations}.

\subsection{Sampling and Positioning the Needle}\label{sec:sampling}

We describe how we sample distractors in $\mathcal{C}$ and how the needle is positioned. As in prior work~\citep{kamradt2023niah, hsieh2024ruler}, varying the needle position isolates positional effects. Our goal is to assess whether increasing distractor density degrades performance even when length-only explanations~\citep{liu2024lost} would not predict a decline.

\noindent\textbf{Distractor sampling.}\quad
For each query, we sample distractors from the candidate pool after excluding the needle. We use a query-specific random seed to fix both the sampled distractor set and its ordering across all experimental conditions. This ensures that performance differences across conditions are not driven by incidental variation in which distractors are sampled.

\noindent\textbf{Needle position.}\quad
We define the needle position by the candidate index $j \in \{1, \dots, |\mathcal{C}|\}$ at which the needle is inserted. Since candidates are concatenated in order and have similar average length (Appendix~\ref{app:candidates_distributions}), $j$ serves as a discrete proxy for the needle's token offset in the prompt. We partition $\mathcal{C}$ into eight contiguous buckets of equal size. This granularity is feasible for all three candidate sizes, $|\mathcal{C}| \in \{200, 400, 4000\}$, and yields buckets large enough to support multiple samples per bucket. To account for randomness, we uniformly sample three insertion indices for each query within each bucket and insert the needle into those indices. 


\subsection{Lexical Density: Moving-Average Type-Token Ratio (MATTR)} \label{sec:density}

To estimate lexical density, we use \emph{lexical diversity}, measured by the Moving-Average Type-Token Ratio (MATTR)~\citep{covington2010mattr}.
Given a token sequence $w_1, \dots, w_N$ and a window size $W$, MATTR averages the proportion of unique tokens over all contiguous windows of length $W$. It is $\mathrm{MATTR}(W) = \frac{1}{N-W+1} \sum_{i=1}^{N-W+1} |\{w_i, \dots, w_{i+W-1}\}|/W$.


We set $W=100$ following common practice in corpus linguistics~\citep{covington2010mattr}.\footnote{Throughout the remainder of the paper, MATTR denotes this variant with window size $W=100$.} MATTR ranges in $[1/W, 1]$: lower values indicate more repetitive text, while higher values indicate greater lexical diversity. We use MATTR as a proxy for lexical density, assuming that greater lexical diversity correlates with more distinct information. While imperfect (e.g., near-synonyms can inflate diversity without adding information), it provides a simple and practical measure for our analysis; we discuss its limitations in Section~\ref{sec:limitations}.

\begin{wraptable}{r}{0.58\linewidth}
\vspace{-2em}
\centering
\small
\setlength{\tabcolsep}{4pt}
\renewcommand{\arraystretch}{1.05}
\caption{\textbf{Lexical-density rankings are robust across metrics.} 
Across MATTR windows, MTLD, entropy, and Yule's K, \textbf{WordChecker is densest}, \textbf{Scene-Rules is intermediate}, and \textbf{MK-NIAH is sparsest}.}
\label{tab:density_metric}

\resizebox{\linewidth}{!}{%
\begin{tabular}{@{}lccc@{}}
\toprule
Metric                      & MK-NIAH    & Scene-Rules & WordChecker \\
\midrule
MATTR $W=50$~[$\uparrow$]        & 0.66       & 0.84        & 1.00        \\
\textbf{MATTR $W=100$~[$\uparrow$]} & \textbf{0.58} & \textbf{0.75} & \textbf{1.00} \\
MATTR $W=200$~[$\uparrow$]       & 0.54       & 0.64        & 1.00        \\
MTLD~[$\uparrow$]            & 37.65      & 113.76      & $>$16k      \\
Shannon Entropy~[$\uparrow$] & 7.81       & 8.79        & 11.97       \\
Yule's K~[$\downarrow$]      & 312.30     & 92.01       & 0.01        \\
\bottomrule
\end{tabular}%
}

\vspace{-1em}
\end{wraptable}


Table~\ref{tab:density_metric} reports multiple lexical diversity metrics (definitions in Appendix~\ref{app:mtld}), all yielding the same ordering: MK-NIAH is the sparsest; Scene-Rules and WordChecker are designed to extend the range toward higher-density settings. The gap is structural: MK-NIAH consists of repeated template text (e.g., \textit{``One of the special magic uuids for \texttt{<uuid1>} is \texttt{<uuid2>}’’}), with variation only in the UUIDs, resulting in low lexical diversity.


\subsection{Synthetic Density Manipulation}\label{sec:manipulation}

The benchmarks of Section~\ref{sec:benchmarks} differ in density, but also in matching mechanism. To establish a link between lexical density and performance, we follow the example of \citet{levy2024same} and control density \emph{within} each benchmark by repeating distractors. For each benchmark, we pick a number $k$ of unique distractors, sample candidates, and repeat them until we reach $|\mathcal{C}|$ items. We then shuffle candidates to introduce randomness, and introduce the needle at the selected bucket position. Smaller $k$ means more repetition and lower MATTR; $k=|\mathcal{C}|$ recovers the original benchmark. 

The needle, its position, the prompt length, and the candidate count all remain the same -- only the number of unique distractors, and therefore the lexical diversity, changes.
We sweep $k$ across the following values: $\{1, 2, 20, 50, 200\}$ for MK-NIAH, $\{1, 2, 40, 100, 400\}$ for Scene-Rules, and $\{1, 20, 400, 1000, 4000\}$ for WordChecker. We chose these values to keep the multiplicative spacing roughly comparable between benchmarks. 
Our manipulation can only reduce lexical density, as it relies on repetition of existing distractors and cannot introduce new lexical content; thus, already sparse benchmarks cannot be densified.

\subsection{Prompt format}

We use the same prompt template (Appendix~\ref{app:examples_benchmarks}) across all benchmarks and conditions, designed to keep the framing natural and avoid steering the model toward any particular solution strategy. Each prompt assigns the model a brief role, describes the input, states the retrieval objective, and requires a structured JSON output containing the predicted answer, a confidence score, and a one-sentence justification. We do not provide in-context examples or step-by-step reasoning instructions; the justification field is included only to support parsing and post-hoc error analysis.

\section{Results}\label{sec:results}
\subsection{Experiment Settings}

\begin{wraptable}{r}{0.52\textwidth}
\vspace{-1.2em}
\centering
\footnotesize
\setlength{\tabcolsep}{2.5pt}
\caption{\textbf{Models evaluated.} The suite covers \textbf{8 recent open LLMs} across \textbf{9B--685B parameters}, mixing \textbf{reasoning} and non-reasoning models, \textbf{dense} and \textbf{MoE} architectures, and both \textbf{local} and API-based inference. MoE sizes are total/active parameters.}
\label{tab:models}
\begin{tabular}{@{}lcccc@{}}
\toprule
Model & Params & R & MoE & Run \\
\midrule
\href{https://arxiv.org/abs/2512.02556}{DeepSeek-V3.2} 
  & 685B & \textbf{Y} & \textbf{Y} & API \\
\href{https://docs.mistral.ai/models/model-cards/mistral-small-4-0-26-03}{Mistral Small 4} 
  & 119B / 6.5B & \textbf{Y} & \textbf{Y} & API \\
\href{https://huggingface.co/meta-llama/Llama-4-Maverick-17B-128E-Instruct}{Llama 4 Maverick} 
  & 400B / 17B & N & \textbf{Y} & API \\
\href{https://arxiv.org/abs/2508.10925}{GPT-OSS-20B} 
  & 21B / 3.6B & \textbf{Y} & \textbf{Y} & Local \\
\href{https://arxiv.org/abs/2508.10925}{GPT-OSS-120B} 
  & 117B / 5.1B & \textbf{Y} & \textbf{Y} & Local \\
\href{https://qwen.ai/blog?id=qwen3.5}{Qwen3.5-9B} 
  & 9B & \textbf{Y} & N & Local \\
\href{https://qwen.ai/blog?id=qwen3.5}{Qwen3.5-27B} 
  & 27B & \textbf{Y} & N & Local \\
\href{https://qwen.ai/blog?id=qwen3.5}{Qwen3.5-397B-A17B} 
  & 397B / 17B & \textbf{Y} & \textbf{Y} & API \\
\bottomrule
\end{tabular}
\vspace{-2em}
\end{wraptable}

We investigate the performance of 8 recent open-source state-of-the-art LLMs, which we list in Table~\ref{tab:models}, ranging from 9 billion to 685 billion parameters. Models come from different families, and include reasoning and non-reasoning variants from both dense and Mixture of Experts (MoE) architectures. Our goal is to assess whether density is a context-confounding factor even for last-generation LLMs.

We employ a hybrid inference strategy: we evaluate the largest architectures through commercial APIs, while we run the remaining models locally on an HPC cluster with NVIDIA A40 and H200 GPUs using vLLM~\citep{kwon2023vllm} in bfloat16 precision. For all models, we set the temperature to T=0.1, ensuring reproducible outputs while preserving some flexibility for multi-step reasoning. We cap output length at 4096 tokens for all models except the Qwen3.5 family where we raise the cap to 16k tokens and apply a mild repetition penalty to mitigate degenerate looping.\footnote{Qwen models frequently enter repetitive reasoning traces and truncate before producing the answer with short limits.} We treat this as a model-specific failure mode: task-accuracy results remain comparable, but cross-family comparisons should be interpreted with this difference in mind.
We evaluate each benchmark on 100 queries; for each query we test 24 needle positions (8 buckets $\times$ 3 samples) and 5 density levels, yielding $100\times24\times5=12{,}000$ inferences per model per benchmark.\footnote{Tables report only mean values for space, but each cell aggregates between 300 and 12{,}000 inferences depending on the level of marginalization (positions, densities). Confidence intervals are shown in Figure~\ref{fig:fig_1} as a representative case.}


\subsection{Positional Decay Under High Density}\label{sec:positional-decay}

We first evaluate the three benchmarks at maximum density ($k=|\mathcal{C}|$), systematically varying the position of the needle. Concretely, we use needle position -- the canonical axis along which long-context performance degrades in prior work -- as a controlled probe with fixed density. We aggregate retrieval scores into common token-position bins (2k, 4k, 6k, 8k, 10k, 12k) for cross-benchmark comparison. Table~\ref{tab:change_position} details the absolute results while Figure~\ref{fig:fig_1} provides the relative visual summary.

\noindent\textbf{The bottom row shows the key finding:}\quad All models degrade as the needle moves deeper into the context, except on MK-NIAH, where performance remains stable. Averaged across the eight models, the drop relative to the 2k baseline at the deepest bin is $+1\%$ on MK-NIAH (MATTR=$0.58$), $-27\%$ on Scene-Rules (MATTR=$0.75$), and $-31\%$ on WordChecker (MATTR=$1.0$). Despite their different matching mechanisms, the three benchmarks show the same trend: higher MATTR (density) produces earlier and more severe positional decay.

{
\footnotesize
\setlength{\tabcolsep}{4pt}
\renewcommand{\arraystretch}{1}
\begin{table}[tb]
\centering
\caption{\textbf{Retrieval Score by needle-position} on MK-NIAH, Scene-Rules, and WordChecker. For cross-benchmark comparability, \textbf{we aggregate results into common token-position bins}. Colors encode accuracy (green = higher, red = lower). \textbf{Bold} indicates the best position for each model. The bottom row reports the \textbf{average drop} across models against the 2k baseline.}
\vspace{4pt}
\resizebox{\textwidth}{!}{%
\begin{tabular}{l cccccc @{\hspace{1.5em}} cccccc @{\hspace{1.5em}} cccccc}
\toprule
& \multicolumn{6}{c}{\textbf{MK-NIAH} ($k = 200$)}
& \multicolumn{6}{c}{\textbf{Scene-Rules} ($k = 400$)}
& \multicolumn{6}{c}{\textbf{WordChecker} ($k = 4000$)} 
 \\
\cmidrule(lr){2-7} \cmidrule(lr){8-13} \cmidrule(l){14-19}
\textbf{Model}
& 2k & 4k & 6k & 8k & 10k & 12k 
& 2k & 4k & 6k & 8k & 10k & 12k 
& 2k & 4k & 6k & 8k & 10k & 12k  \\
\midrule
DeepSeek-V3.2
& \h{0.99} & \hb{0.99} & \h{0.97} & \h{0.97} & \h{0.98} & \h{0.98}
& \hb{0.98} & \h{0.85} & \h{0.78} & \h{0.66} & \h{0.62} & \h{0.63} 
& \hb{0.86} & \h{0.64} & \h{0.57} & \h{0.54} & \h{0.39} & \h{0.47} \\
Mistral Small 4
& \h{0.98} & \hb{1.00} & \hb{1.00} & \h{0.99} & \h{0.99} & \h{0.99} 
& \hb{0.86} & \h{0.66} & \h{0.52} & \h{0.36} & \h{0.31} & \h{0.34} 
& \hb{0.72} & \h{0.56} & \h{0.36} & \h{0.21} & \h{0.17} & \h{0.23} \\
LLama 4 Maverick
& \h{0.99} & \h{0.99} & \h{0.97} & \hb{1.00} & \h{0.93} & \h{0.96} 
& \hb{0.95} & \h{0.85} & \h{0.73} & \h{0.58} & \h{0.55} & \h{0.59} 
& \hb{0.88} & \h{0.74} & \h{0.52} & \h{0.39} & \h{0.33} & \h{0.42} \\
GPT-OSS-20B
& \h{0.99} & \hb{1.00} & \hb{1.00} & \hb{1.00} & \hb{1.00} & \hb{1.00}
& \hb{0.82} & \h{0.74} & \h{0.67} & \h{0.59} & \h{0.56} & \h{0.62} 
& \hb{0.91} & \h{0.80} & \h{0.71} & \h{0.46} & \h{0.59} & \h{0.34}  \\
GPT-OSS-120B 
& \hb{0.95} & \h{0.90} & \h{0.88} & \h{0.96} & \h{0.92} & \h{0.84}
& \hb{0.95} & \h{0.83} & \h{0.79} & \h{0.77} & \h{0.73} & \h{0.62} 
& \hb{0.96} & \h{0.89} & \h{0.77} & \h{0.63} & \h{0.73} & \h{0.41} \\
Qwen3.5-9B 
& \hb{1.00} & \hb{1.00} & \hb{1.00} & \hb{1.00} & \hb{1.00} & \hb{1.00}
& \hb{0.95} & \h{0.87} & \h{0.85} & \h{0.81} & \h{0.82} & \h{0.82} 
& \h{0.44} & \h{0.34} & \h{0.35} & \h{0.36} & \h{0.43} & \hb{0.46} \\
Qwen3.5-27B
& \h{0.90} & \hb{0.97} & \h{0.96} & \h{0.96} & \h{0.93} & \h{0.94}
& \hb{0.98} & \h{0.93} & \h{0.89} & \h{0.83} & \h{0.86} & \h{0.83} 
& \h{0.79} & \h{0.72} & \h{0.79} & \hb{0.83} & \h{0.75} & \h{0.76} \\
Qwen3.5-397B
& \hb{1.00} & \hb{1.00} & \hb{1.00} & \hb{1.00} & \hb{1.00} & \h{0.99}
& \hb{0.94} & \h{0.82} & \h{0.86} & \h{0.82} & \h{0.82} & \h{0.85} 
& \h{0.96} & \h{0.95} & \h{0.94} & \hb{0.98} & \h{0.92} & \h{0.93} \\
\midrule
\textbf{Avg.\ $\Delta$ vs.\ 2k}
& --- & \posd{$+0.01$} & \negd{$-0.00$} & \posd{$+0.01$} & \posd{$+0.01$} & \posd{$+0.01$}
& --- & \negd{$-0.11$} & \negd{$-0.17$} & \negd{$-0.25$} & \negd{$-0.27$} & \negd{$-0.27$}
& --- & \negd{$-0.11$} & \negd{$-0.19$} & \negd{$-0.27$} & \negd{$-0.28$} & \negd{$-0.31$} \\
\bottomrule
\end{tabular}%
}
\label{tab:change_position}
\vspace{-1.0em}
\end{table}
}

\noindent\textbf{MK-NIAH saturates.}\quad Having the \textbf{lowest density}, retrieval scores are near-perfect across all positions, with no clear trend as the needle moves -- consistent with~\citet{hsieh2024ruler}. At $\sim$12k tokens, MK-NIAH is too sparse to challenge any model, including the smaller ones; positional decay simply does not exist in this regime.

\noindent\textbf{Scene-Rules breaks the position axis.}\quad With \textbf{medium density}, moving the needle from the first to the second bin (2--4k tokens) already produces an average $11.1\%$ relative drop, and performance keeps decaying until plateauing around 10k. The drop is heterogeneous across models: Mistral Small~4, despite matching GPT-OSS-120B in parameter count and MoE architecture, loses $61\%$ relative between 2k and 12k. Qwen models, by contrast, generate substantially longer outputs (~3k tokens vs. ~87 for others), which may help mitigate performance collapse under high-density conditions~\citep{zhu-etal-2025-chain}; even so, every Qwen variant loses $\sim$10 points between 2k and 4k. 

\noindent\textbf{WordChecker is semantically simple but search-intensive.}\quad
The target is a forbidden word whose lemma appears in the query, requiring no semantic inference -- yet WordChecker yields the largest average drop of the three benchmarks ($-31\%$ at 12k). Five of eight models lose 45--68\% of their 2k accuracy by 12k, including large MoE systems such as GPT-OSS-120B ($0.96 \to 0.41$) and DeepSeek-V3.2 ($0.86 \to 0.47$); only the larger Qwen variants (27B, 397B) remain stable. This shows that long-context degradation is not driven only by reasoning difficulty: even matching a lemma against a list -- about as simple as retrieval gets -- collapses once the list becomes dense enough.

\begin{table}[t]
\centering
\caption{\textbf{Density sweep at fixed length and position structure.}
Lexical density is \textbf{reduced} left-to-right via the number of unique
keys / rules / words: the leftmost column is the original benchmark of
Section~\ref{sec:positional-decay}, the rightmost the most sparsified variant. Each
cell averages retrieval score across needle positions. Colour encodes accuracy
(green = high, red = low); \textbf{bold} marks the best density per model.
The bottom row reports the \textbf{average change} across models relative to
the \textbf{max-density baseline} (leftmost column) within each benchmark.}
\label{tab:density_sweap}
\setlength{\tabcolsep}{5pt}
\renewcommand{\arraystretch}{1.15}
\resizebox{\textwidth}{!}{%
\footnotesize
\begin{tabular}{@{}l @{\hskip 10pt} ccccc @{\hskip 12pt} ccccc @{\hskip 12pt} ccccc@{}}
\toprule
\multirow{2}{*}{\textbf{Model}}
 & \multicolumn{5}{c}{\textbf{MK-NIAH} \textit{(\# Unique Keys)}}
 & \multicolumn{5}{c}{\textbf{Scene-Rules} \textit{(\# Unique Rules)}}
 & \multicolumn{5}{c}{\textbf{WordChecker} \textit{(\# Unique Words)}} \\
\cmidrule(lr){2-6} \cmidrule(lr){7-11} \cmidrule(lr){12-16}
 & 200 & 50 & 20 & 2 & 1
   & 400 & 100 & 40 & 2 & 1
   & 4k & 1k & 400 & 20 & 1 \\
\midrule
DeepSeek-V3.2    & \h{0.98}  & \h{0.98}  & \h{0.98}  & \h{0.98}  & \hb{0.99} & \h{0.75}  & \h{0.82}  & \h{0.90}  & \hb{0.98} & \hb{0.98} & \h{0.58}  & \h{0.47}  & \h{0.57}  & \h{0.88}  & \hb{0.98} \\
Mistral Small 4  & \h{0.99}  & \h{0.99}  & \hb{1.00} & \h{0.99}  & \h{0.99}  & \h{0.51}  & \h{0.52}  & \h{0.66}  & \h{0.99}  & \hb{1.00} & \h{0.38}  & \h{0.27}  & \h{0.29}  & \h{0.57}  & \hb{0.98} \\
Llama 4 Maverick & \h{0.96}  & \h{0.97}  & \hb{0.99} & \h{0.97}  & \h{0.87}  & \h{0.71}  & \h{0.66}  & \h{0.69}  & \hb{0.96} & \h{0.91}  & \h{0.55}  & \h{0.30}  & \h{0.20}  & \h{0.18}  & \hb{0.91} \\
GPT-OSS-20B      & \hb{1.00} & \h{0.99}  & \hb{1.00} & \hb{1.00} & \h{0.98}  & \h{0.66}  & \h{0.71}  & \h{0.82}  & \h{0.93}  & \hb{0.97} & \h{0.63}  & \h{0.57}  & \h{0.63}  & \h{0.82}  & \hb{0.94} \\
GPT-OSS-120B     & \h{0.91}  & \h{0.96}  & \hb{0.99} & \h{0.93}  & \h{0.47}  & \h{0.78}  & \h{0.88}  & \h{0.95}  & \hb{1.00} & \hb{1.00} & \h{0.73}  & \h{0.69}  & \h{0.71}  & \h{0.86}  & \hb{0.97} \\
Qwen3.5-9B       & \hb{1.00} & \hb{1.00} & \h{0.99}  & \hb{1.00} & \h{0.99}  & \h{0.85}  & \h{0.90}  & \h{0.94}  & \hb{0.99} & \h{0.94}  & \h{0.40}  & \h{0.07}  & \h{0.07}  & \h{0.28}  & \hb{0.47} \\
Qwen3.5-27B      & \h{0.94}  & \h{0.97}  & \h{0.96}  & \hb{0.97} & \h{0.95}  & \h{0.89}  & \h{0.96}  & \h{0.97}  & \hb{0.99} & \h{0.94}  & \h{0.77} & \h{0.47}  & \h{0.78}  & \h{0.73}  & \hb{0.86}  \\
\midrule
\textbf{Avg.\ $\Delta$ vs.\ max-density}
& --- & \posd{$+0.01$} & \posd{$+0.02$} & \posd{$+0.01$} & \negd{$-0.08$}
& --- & \posd{$+0.04$} & \posd{$+0.11$} & \posd{$+0.24$} & \posd{$+0.23$}
& --- & \negd{$-0.17$} & \negd{$-0.11$} & \posd{$+0.04$} & \posd{$+0.30$} \\
\bottomrule
\end{tabular}%
}
\vspace{-1.3em}
\end{table}


\subsection{Decoupling Lexical Density from Task Complexity} \label{sec:decoupling}

We established that denser benchmarks break down earlier than sparser ones at fixed length. A residual concern is that the three benchmarks differ not only in lexical density but also in task and content. To rule out task complexity as the driver, we apply the within-benchmark sparsification described in Section~\ref{sec:methodology}: task, length, and needle position are held fixed; only the number of unique distractors $k$ (thus density) changes. Table~\ref{tab:density_sweap} reads max-density on the left (the original benchmark of Section~\ref{sec:positional-decay}) and progressively sparser variants to the right; the bottom row reports the average score change across models relative to this given-density baseline.

\noindent\textbf{MK-NIAH: ceiling, with one curious artefact.}\quad MK-NIAH already sits at the sparse end of the density spectrum, so further sparsification cannot make it meaningfully easier. Models stay at ceiling for $k\geq 2$ ($|\Delta|\leq 0.02$). The only outlier appears at $k{=}1$, where the average drops by $-8\%$ -- driven almost entirely by GPT-OSS-120B ($-47\%$). Trace inspection shows that, with a single repeated UUID, the task becomes \emph{too} easy for it: the model parses the correct identifier but returns only its numeric characters, contrary to the prompt instructions. We treat this as a degenerate-output failure rather than a density effect.

\noindent\textbf{Scene-Rules: clean monotonic recovery.}\quad As we sparsify the haystack, average accuracy climbs monotonically: $+4\%$, $+11\%$, $+24\%$, $+23\%$ relative to the max-density baseline. Only lexical density changes, and each step toward sparser inputs systematically restores performance. In past works, length- and position-only frameworks would predict no change in our regime. Instead, we observe a $24\%$ accuracy swing.

\begin{wrapfigure}{r}{0.55\linewidth}
    \centering
    \vspace{-1em}
    \includegraphics[width=\linewidth]{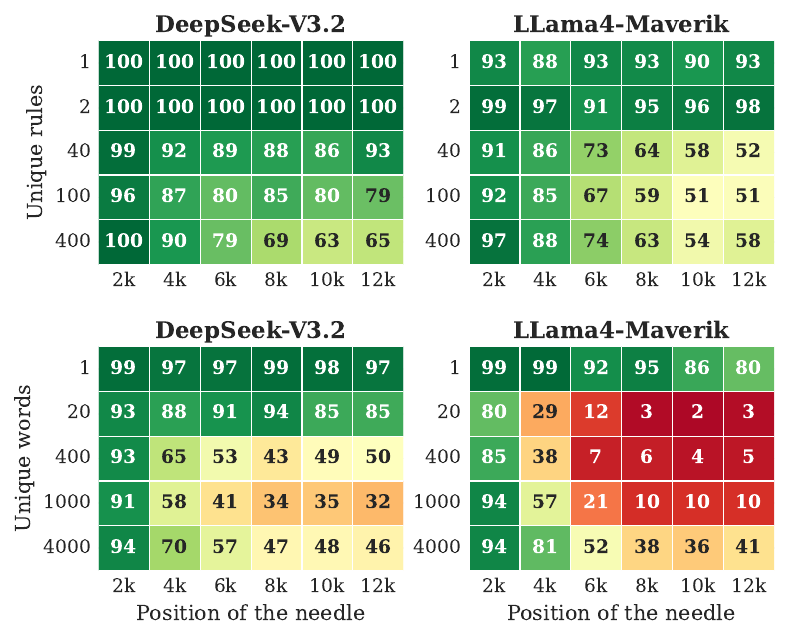}
    \caption{\textbf{Density and position interact} on Scene-Rules (top) and WordChecker (bottom). Low density flattens positional decay; high density activates it.}
    \label{fig:density_position}
    \vspace{-1em}
\end{wrapfigure}

Figure~\ref{fig:density_position} (top) makes the joint view explicit: at
the lowest density, accuracy is flat across all needle positions; positional
decay emerges only as density climbs. Density and position do not act
independently -- density is what \emph{turns on} the position effect.

\noindent\textbf{WordChecker: noisier, but the density signal holds.}\quad On WordChecker the recovery is non-monotonic: at moderate sparsification (unique distractors $k{=}1\text{k}, 400$) accuracy is actually \emph{worse} than at the max-density baseline ($-17\%$, $-11\%$), and recovers only at $k{=}20$ and $k{=}1$ ($+4\%$, $+30\%$).
Trace inspection identifies the cause. Several models adopt a two-step heuristic -- deduplicate the candidate list, then iterate to find a match.

With one or two repeated distractors plus the needle, the heuristic succeeds; with heavier repetition, the deduplication
step strips the needle too, after which models either abstain, fall into infinite-iteration truncation, or return a word semantically associated with the sentence but absent from the candidate list (e.g., \texttt{hike} for a sentence about skiing in the mountains). The non-monotonicity is therefore an artefact of how synthetic sparsification interacts with this strategy: the max-density baseline ($k=4\text{k}$) still produces the worst absolute scores observed in the paper, while the maximally sparse variant ($k=1$) recovers $+30\%$ on average. Only Qwen3.5-9B fails due to its limited size -- as we later discuss. We cover cleaner sparsification protocols in Section~\ref{sec:limitations}.

Figure~\ref{fig:density_position} (bottom) provides the details: for Llama-4 Maverick, density $k{=}20$ is the worst across nearly all positions, while $k{=}1$ recovers ceiling performance; for DeepSeek-V3.2, the same non-monotonicity appears attenuated, with $k{=}1\text{k}$ underperforming the $k{=}4\text{k}$ baseline at deep positions.

\begin{figure}[h!]
    \centering
    \begin{subfigure}[b]{0.4\linewidth}
        \centering
        \includegraphics[width=\linewidth]{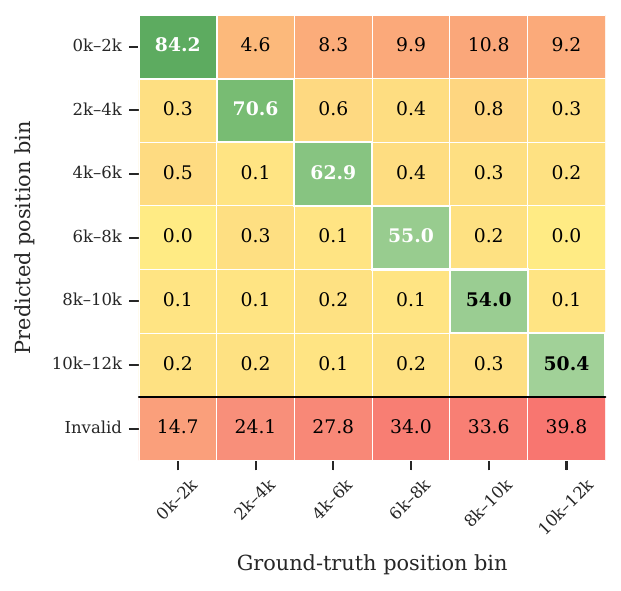}
        \caption{WordChecker -- Position of the needle: \textbf{predictions vs.\ ground truth}. Average of 8 models.}
        \label{fig:wordchecker_pred_vs_gt}
    \end{subfigure}
    \hfill
    \begin{subfigure}[b]{0.48\linewidth}
        \centering
        \includegraphics[width=\linewidth]{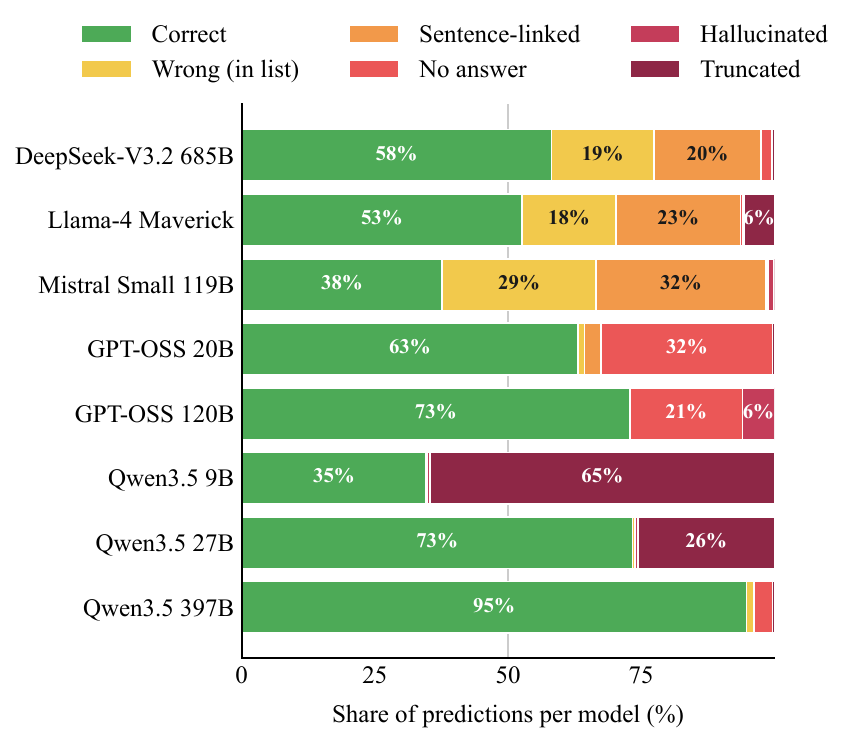}
        \caption{WordChecker -- \textbf{Breakdown of models' predictions}, averaged over bins.}
        \label{fig:wordchecker_breakdown}
    \end{subfigure}

    \caption{\textbf{WordChecker error analysis.} Predicted vs.\ ground truth needle positions (left) show that higher density biases model attention toward earlier context rather than uniformly degrading it; decomposing predictions by outcome (right) reveals three qualitatively distinct failure regimes: conflation, abstention, and loop inversion.}
    \label{fig:error_analysis}
\end{figure}

\subsection{Deconstructing the Breakdown}\label{sec:deconstructing}

Aggregate scores tell us \emph{that} density triggers earlier collapse; they do not tell us \emph{how}. We use WordChecker -- the maximum-density,
minimum-reasoning case -- as a clean playground to investigate failure modes.

\noindent\textbf{Where models look (Figure~\ref{fig:wordchecker_pred_vs_gt}).}\quad We compare predicted to true needle positions, averaged across models. Two effects emerge as the needle moves deeper: i) models increasingly retrieve content from the \emph{earliest} bins rather than from the correct one (top row); and ii) the share of invalid outputs grows with depth (bottom row). This suggests that density does not uniformly degrade attention, but biases it toward the beginning of the context.

\noindent\textbf{How models fail (Figure~\ref{fig:wordchecker_breakdown}).}\quad Decomposing the predictions of each model by outcome reveals three different
behavioural regimes. \textbf{(i) Conflation} (Mistral Small, Llama-4 Maverick, DeepSeek-V3.2): probability mass splits between correct answers, wrong words from the candidate list, and \emph{sentence-linked false alarms} -- declaring a forbidden word present but returning one from the sentence under validation. Both types of error grow with needle depth (Appendix~\ref{app:errors}): under high density, these models conflate the two textual inputs. \textbf{(ii) Abstention} (GPT-OSS-20B, GPT-OSS-120B): both variants increasingly refuse to answer, with abstention concentrated in deeper bins. Density does not push these models toward wrong predictions -- it pushes them out of the task. \textbf{(iii) Loop inversion} (Qwen3.5 family): reasoning traces show that the larger Qwen variants walk through sentence words and check each against the candidate list, rather than the reverse. The two procedures are logically equivalent, but keeping the long collection as the inner loop is decisive at high density: Qwen3.5-397B and Qwen3.5-27B reach $95\%$ and $73\%$ accuracy, while Qwen3.5-9B attempts the same strategy but truncates in $\sim$65\% of cases.

The observed breakdown is not simply a stronger version of the same error as density increases. Instead, it reflects a regime change: under high-density inputs, different model families exhibit qualitatively distinct failure modes, revealing how their architectures and inference strategies cope when no tokens can be ignored.

\noindent\textbf{Takeaway.}\quad On $2/3$ benchmarks, reducing lexical density at fixed length,
position, and task structure produces large accuracy gains; on the third it
cannot, only because the benchmark is already saturated at the sparse end.
Figure~\ref{fig:density_position} sharpens the picture: density does not
merely lower mean accuracy -- it activates and amplifies the positional
decay observed in Section~\ref{sec:positional-decay}. Together with the controlled position sweep, these results show that lexical density is a causal contributor to long-context breakdown, interacting with position in ways that length-only explanations cannot capture.

\vspace{-0.5em}
\section{Discussion and Limitations}\label{sec:limitations}

Overall, our results show that effective context capacity depends not only on token budget and needle position, but also on the lexical density of the surrounding context. While our benchmarks enable controlled analysis of lexical density, this work has implications and limitations that define its scope and suggest directions for future research.

\noindent\textbf{Implications for compression-based approaches.}\quad
Many recent methods reduce context length via prompt or memory compression, but in doing so, increase lexical density. Our results highlight a potential trade-off: denser contexts can degrade retrieval performance even at moderate lengths. This raises the question of whether compression strategies, while efficient in token usage, may inadvertently reduce context reliability.

\noindent\textbf{Mechanistic understanding.}\quad
We characterize the behavioural impact of lexical density but do not provide a mechanistic explanation of its origin. One plausible hypothesis is that attention has limited effective resolution: in low-density contexts, redundancy allows compression over repeated spans, whereas in high-density contexts each token introduces distinct information, forcing attention to distribute more broadly and reducing retrieval precision. Preliminary results in Appendix~\ref{app:post_target_ablation} show that truncating context after the needle barely improves retrieval, suggesting that distance tokens do not contribute to the ``density penalty''. We leave mechanistic validation to future work.

\noindent\textbf{Limits of density manipulation.}\quad Our within-benchmark intervention varies density by repeating existing distractors, which can only reduce lexical diversity and cannot generate higher-density contexts. As a result, high-density regimes are studied through cross-benchmark comparisons, while within-benchmark analyses are limited to sparsification. Designing controlled methods to increase density without altering task structure remains challenging and shall be studied in future work.

\noindent\textbf{Measurement of density.}\quad We measure lexical density using the Moving-Average Type–Token Ratio (MATTR), a proxy for lexical diversity. Although correlated with information load, MATTR does not capture semantic redundancy: near-synonyms or paraphrases may increase diversity without adding new information. Our findings should therefore be interpreted as effects of lexical diversity, not a complete characterization of information-theoretic density. Developing more principled measures, potentially grounded in information theory, remains an open problem.

\noindent\textbf{Synthetic evaluation setting.}\quad Our experiments rely on controlled benchmarks designed to isolate density effects. While this enables causal analysis, real-world contexts often exhibit additional structure -- such as topical coherence, hierarchy, or redundancy patterns -- that may interact with density in ways not captured here. Preliminary evidence suggests that many practical inputs (e.g., agent configurations or technical documentation) already operate in high-density regimes, but validating these effects in naturalistic settings remains necessary.

\noindent\textbf{Scope of contribution.}\quad This work is diagnostic: we identify and isolate a previously overlooked factor in long-context degradation, but do not propose mitigation strategies. Developing methods to improve robustness under high-density inputs is a natural next step.



\bibliographystyle{plainnat}
\bibliography{main}
\appendix
\newpage

\section{Benchmark details: evaluation prompts and dataset construction}\label{app:examples_benchmarks}

This section explains the empirical framework used in our study, connecting the formal method described in Section \ref{sec:methodology} with the practical steps needed to evaluate the model. It is divided into three main parts: an analysis of cognitive load and qualitative prompt structures (Section \ref{app:qualitative_prompts}), the exact system instructions and structured output schemas used to ensure reliable parsing (Section \ref{app:full_prompts_schemas}), and the multi-stage synthetic generation pipeline used to build the Scene-Rule dataset (Section \ref{app:scene_rule_dataset}).

\subsection{Qualitative examples and cognitive profiles}
\label{app:qualitative_prompts}
This subsection gives examples of the prompt formats tested in our study. For each benchmark, we show the query $q$, a sample of the context $\mathcal{C}$, and the expected answer $y^\star$. We also include a cognitive profile that explains the retrieval process in simple reasoning steps, from basic keyword matching to more complex constraint-based reasoning.

\paragraph{Multi-Key Needle-in-a-Haystack (MK-NIAH).}
MK-NIAH is our low-density baseline. It tests exact key-value retrieval in highly noisy contexts. The use of artificial UUIDs prevents semantic shortcuts, so the model must rely only on literal string matching. The prompt contains repetitive boilerplate text with many similar UUID pairs acting as distractors.

\begin{tcolorbox}[colback=blue!5!white, colframe=blue!70!black, boxrule=0.5pt, arc=2pt, title=Qualitative prompt: MK-NIAH]
\textbf{Query ($q$):} \\
What is the special magic uuid for \textcolor{blue}{\texttt{f488e328-[...]9f}} mentioned in the following text?

\vspace{2mm}
\textbf{Context ($\mathcal{C}$):} \\
One of the special magic uuids for \texttt{36089374-[...]81} is: \texttt{602ba256-[...]74}.\\
One of the special magic uuids for \textbf{\texttt{f488e328-[...]9f}} is: \textcolor{red}{\texttt{7c65a804-[...]c2}}.\\
One of the special magic uuids for \texttt{9ab8180f-[...]d5} is: \texttt{ae8e6075-[...]3c}. \\
One of the special magic uuids for \texttt{acb8480f-[...]d5} is: \texttt{a034d075-[...]19}.

\vspace{2mm}
\textbf{Target answer ($y^\star$):} \texttt{7c65a804-[...]c2}
\end{tcolorbox}

The model must ignore repetitive noise and find the UUID that exactly matches the query. No semantic understanding or reasoning is needed; success mainly reflects literal retrieval and positional recall over long contexts.

\paragraph{Scene-Rule.}
The Scene-Rule benchmark tests semantic understanding and deductive constraint matching in a structured setting. Unlike MK-NIAH, it prevents exact string matching. The model must match a colloquial scene to one of 400 distinct rules written in a strict, pseudo-legal style. The rules act as similar distractors, so success requires semantic alignment rather than surface-level word overlap.

\begin{tcolorbox}[colback=green!2!white, colframe=green!40!black, boxrule=0.5pt, arc=2pt, title=Qualitative prompt: Scene-Rule]
\textbf{Query ($q$):} \\
\textit{Scene:} Jake is experiencing severe pain from a wisdom tooth removal. Despite reading the recommended dosage on his medicine, \textcolor{blue}{he intentionally takes double the amount}.

\vspace{2mm}
\textbf{Context ($\mathcal{C}$):} \\
\texttt{[Id\_234]} It is prohibited to inquire about another individual's medication or medical prescriptions unless explicitly invited to do so.\\
\textcolor{red}{\texttt{[Id\_343]}} Individuals must \textcolor{red}{not consume more than the prescribed dosage of pain medication}, as indicated by a licensed healthcare provider or the medication's packaging.\\
\texttt{[Id\_404]} It is prohibited for an individual to intentionally expose themselves to known sources of pathogens or contaminants with the purpose of inducing illness.\\
\texttt{[Id\_416]} It is prohibited to reference one's illness in a manner that seeks to gain an advantage or manipulate others during a disagreement.

\vspace{2mm}
\textbf{Target answer ($y^\star$):} \texttt{[Id\_343]}
\end{tcolorbox}

Here, the model must connect Jake's action -- intentionally taking more medicine than recommended -- to Rule 343. It must also reject nearby health-related distractors, such as asking about another person's medication or using illness to manipulate an argument.

\textbf{WordChecker.}
WordChecker tests the effect of extreme lexical density ($\text{MATTR} = 1.0$) without requiring complex reasoning. It is a lemma-aware set intersection task: the model must match words in the query to their root forms in the context. Unlike MK-NIAH, this requires partial keyword mapping rather than exact string matching. The context contains up to 4,000 unique, unstructured word distractors.

\begin{tcolorbox}[colback=orange!5!white, colframe=orange!70!black, boxrule=0.5pt, arc=2pt, title=\textbf{Qualitative prompt: WordChecker}]
\textbf{Query ($q$):} \\
\textit{Sentence:} The students sit in a circle as the teacher \textcolor{blue}{talks}.

\vspace{2mm}
\textbf{Context ($\mathcal{C}$):} \\
\texttt{discard - delete - lunch - bib - shutter - shading - sterile - cathedral - secretary - scrub - energy - polo - \textcolor{red}{talk} - bicycle - tradition - computer - friday - skylight - lynx - martini - [...]}

\vspace{2mm}
\textbf{Target answer ($y^\star$):} \texttt{talk}
\end{tcolorbox}

Here, the model must keep the query in memory while scanning a dense list of unique words. It must map the surface form ``talks'' to the lemma ``talk'' and ignore thousands of unrelated distractors.


\subsection{Complete evaluation prompts and structured output schemas}
\label{app:full_prompts_schemas}

To make evaluation deterministic and easy to parse automatically, we require all models to return structured JSON outputs. Each model receives a system prompt containing a serialized Pydantic\footnote{\url{https://docs.pydantic.dev/}} schema. The output must include the final answer, a \texttt{reasoning} field, and a \texttt{confidence} score. The \texttt{reasoning} field is included only as a short justification to support parsing and post-hoc error analysis, not to provide in-context examples or explicitly encourage step-by-step reasoning.

\paragraph{MK-NIAH System Prompt.}
For MK-NIAH, the instruction is intentionally simple. The model is given a brief ``detective'' role to encourage careful extraction. The output schema requires the target UUID under the \texttt{number} key, together with the standard \texttt{confidence} and \texttt{reasoning} fields used for analysis.

\begin{tcolorbox}[colback=blue!5!white, colframe=blue!70!black, boxrule=0.5pt, arc=2pt, title=System prompt template: MK-NIAH]
\small






You are a detective trying to find a secret number hidden within a text. 

\vspace{2mm}
\noindent \textbf{You will be given:}\\
A block of text containing the hidden number.

\vspace{2mm}
\noindent \textbf{YOUR TASK:}\\
Scan the provided text, locate the hidden target number, and extract it exactly as it appears.

\vspace{2mm}
\noindent \textbf{You MUST always output:}\\
- A one-sentence explanation of where the number was found in the text.\\
- The extracted secret number as a string.\\
- A confidence score between 0.0 and 1.0.

\vspace{2mm}
\texttt{\{\{ input\_text \}\}}

\vspace{2mm}
\noindent \textbf{You MUST respond ONLY with a valid JSON object matching this exact schema:}\\
\texttt{\{}\\
\texttt{\hspace*{4mm}"number": string, \hspace*{4mm}\textcolor{gray}{\textit{// The hidden special number.}}}\\
\texttt{\hspace*{4mm}"confidence": float, \hspace*{2.5mm}\textcolor{gray}{\textit{// Confidence score between 0.0 and 1.0.}}}\\
\texttt{\hspace*{4mm}"reasoning": string \hspace*{3.5mm}\textcolor{gray}{\textit{// One-sentence explanation of the reasoning.}}}\\
\texttt{\}}
\end{tcolorbox}

\paragraph{Scene-Rule System Prompt.}
For Scene-Rule, the prompt moves from string extraction to constraint matching. The model is framed as an ``expert legal adjudicator'' to encourage careful rule evaluation. It must compare the scenario against the candidate rules and return the exact violated rule ID (\texttt{conflicting\_rule\_id}). The \texttt{reasoning} field provides a short justification for the selected ID, supporting later error analysis.

\begin{tcolorbox}[colback=green!2!white, colframe=green!40!black, boxrule=0.5pt, arc=2pt, title=System prompt template: Scene-Rule]
\small
You are an expert legal adjudicator and AI compliance judge.

\vspace{2mm}
\noindent \textbf{You will be given:}\\
1. A set of candidate RULES, each with a unique ID and a detailed description.\\
2. A SCENARIO that describes a specific situation or action. The scenario always violates one of the candidate rules.

\vspace{2mm}
\noindent \textbf{YOUR TASK:}\\
For each scenario, analyze the associated candidate rules and determine which rule the scenario violates.

\vspace{2mm}
\noindent \textbf{You MUST always output:}\\
- The ID of the broken rule.\\
- A confidence score between 0 and 1, where 0 means no confidence at all and 1 means absolute certainty.\\
- A one-sentence explanation of why you believe the scenario violates the identified rule.

\vspace{2mm}
\texttt{\{\{ full\_prompt\_text \}\}}

\vspace{2mm}
\noindent \textbf{You MUST respond ONLY with a valid JSON object matching this exact schema:}\\
\texttt{\{}\\
\texttt{\hspace*{4mm}"reasoning": string, \hspace*{16mm}\textcolor{gray}{\textit{// One sentence explaining the violation.}}}\\
\texttt{\hspace*{4mm}"conflicting\_rule\_id": string, \hspace*{2mm}\textcolor{gray}{\textit{// Exact ID of the rule (e.g., 'Id\_42').}}}\\
\texttt{\hspace*{4mm}"confidence": float \hspace*{17.5mm}\textcolor{gray}{\textit{// Confidence score between 0.0 and 1.0.}}}\\
\texttt{\}}
\end{tcolorbox}

\paragraph{WordChecker System Prompt.}
In WordChecker the model is framed as an ``expert compliance analyst'' and must find a forbidden word, or its lemma, from a short sentence within a distractor list of up to 4,000 words. The schema requires the exact \texttt{word}, ensuring the model resolves the morphological match between the sentence and the vocabulary list.

\begin{tcolorbox}[colback=orange!5!white, colframe=orange!70!black, boxrule=0.5pt, arc=2pt, title=\textbf{System Prompt Template: WordChecker}]
\small
You are an expert compliance analyst specializing in rule adherence evaluation.

\vspace{2mm}
\noindent \textbf{You will be given:}\\
1. A SENTENCE \\
2. A list of candidate WORDS

\vspace{2mm}
\noindent \textbf{YOUR TASK:}\\
Identify the single word from the list that is contained within the sentence. \\
We say that a word is "contained" within the sentence if it appears as it is or as a lemma (e.g., "run" is contained in "running").

\vspace{2mm}
\noindent \textbf{You MUST always output:}\\
- The candidate words that the sentence contains.\\
- A confidence score between 0 and 1, where 0 means no confidence at all and 1 means absolute certainty.\\
- A one-sentence explanation of why you believe the scenario violates the identified rule.

\vspace{2mm}
\texttt{\{\{ full\_prompt\_text \}\}}

\vspace{2mm}
\textbf{You MUST respond ONLY with a valid JSON object matching this exact schema:}\\
\texttt{\{}\\
\texttt{\hspace*{4mm}"word": string, \hspace*{11.5mm}\textcolor{gray}{\textit{// The candidate word contained in the sentence.}}}\\
\texttt{\hspace*{4mm}"confidence": float, \hspace*{2.5mm}\textcolor{gray}{\textit{// Confidence score between 0.0 and 1.0.}}}\\
\texttt{\hspace*{4mm}"reasoning": string \hspace*{4.5mm}\textcolor{gray}{\textit{// One-sentence explanation of the reasoning.}}}\\
\texttt{\}}
\end{tcolorbox}

\subsection{Scene-Rule dataset construction}
\label{app:scene_rule_dataset}

To evaluate deductive constraint satisfaction while reducing contamination and ambiguity, we build the Scene-Rule benchmark through a four-step synthetic generation pipeline:

\begin{enumerate}
    \item \textbf{Source selection and diversity sampling.} 
    We start from the public test split of the Moral Stories dataset \citep{emelin2020moralstoriessituatedreasoning}. We manually remove instances containing vague or highly subjective terms, then use K-means clustering to sample diverse seed examples from the filtered set.

    \item \textbf{Synthetic triplet generation.}
    Using these seeds, we prompt \texttt{gpt-4o} to generate triplets consisting of an objective rule, a violating scene, and a hard-negative non-violating scene. The rules are written as strict, pseudo-legal statements, avoiding subjective language and focusing on observable actions. The scenes are short narratives, with the hard negative designed to closely match the violating scene in wording and structure while not breaking the rule.

    \item \textbf{Automated differential self-verification.} 
    A separate \texttt{gpt-4o} instance verifies each triplet as a literal legal adjudicator. It evaluates both scenes against the same rule and retains the triplet only if it marks the violating scene as a breach (\textit{True}) and the hard-negative scene as non-violating (\textit{False}). This differential check helps ensure that success depends on semantic understanding rather than keyword overlap.

    \item \textbf{Final benchmark assembly.} 
    Triplets that fail this verification step are discarded as ambiguous. The remaining rules and scenes are then used to build the final high-density retrieval contexts, as described in Section \ref{sec:methodology}.
\end{enumerate}

\section{Candidate Distributions}\label{app:candidates_distributions}

This section checks whether distractor candidates have consistent lengths within each benchmark. This matters because our positional analysis uses the candidate index $j$ as a proxy for the token position of the inserted needle, which is valid only when candidate lengths are relatively uniform.

\begin{figure}[h!]
    \centering
    \begin{subfigure}[b]{0.325\linewidth}
        \centering
        \includegraphics[width=\linewidth]{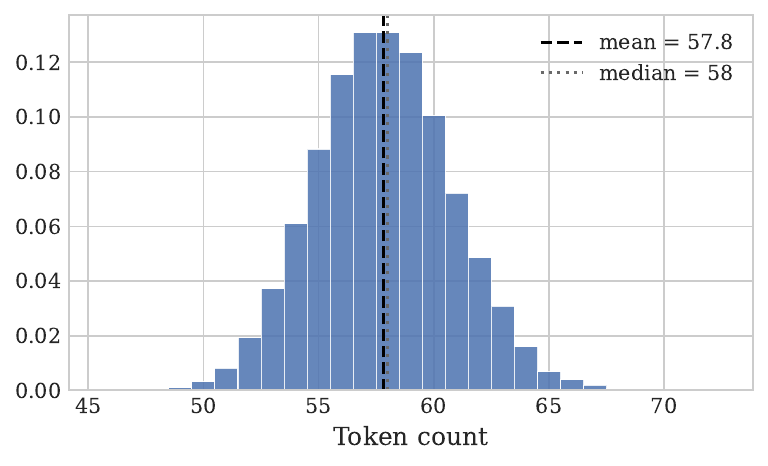}
        \caption{\textit{MK-NIAH}}
    \end{subfigure}
    \hfill
    \begin{subfigure}[b]{0.325\linewidth}
        \centering
        \includegraphics[width=\linewidth]{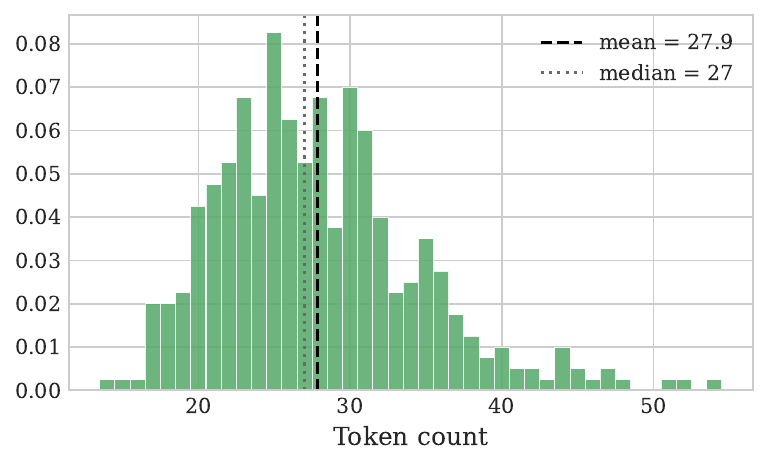}
        \caption{\textit{Scene-Rule}}
    \end{subfigure}
    \hfill
    \begin{subfigure}[b]{0.325\linewidth}
        \centering
        \includegraphics[width=\linewidth]{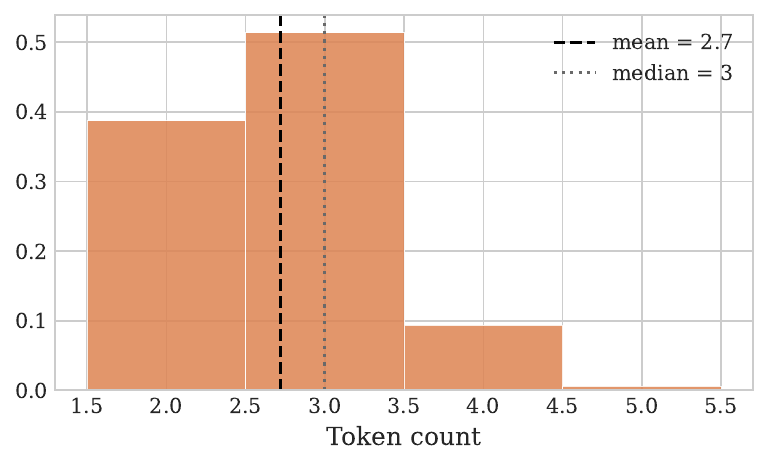}
        \caption{\textit{WordChecker}}
    \end{subfigure}
    
    \caption{\textbf{Candidate token-length distributions.} Histograms show token counts for individual candidates in each benchmark. \textit{MK-NIAH} candidates average 57.8 tokens, \textit{Scene-Rule} candidates average 27.9 tokens, and \textit{WordChecker} candidates average 2.7 tokens. Since lengths are tightly concentrated within each benchmark, the candidate index $j$ is a reliable proxy for absolute token position.}
    \label{fig:token_distributions}
\end{figure}

Figure~\ref{fig:token_distributions} shows the token-length distributions for the three benchmarks. \textit{MK-NIAH} has the longest candidates, with a mean of 57.8 tokens and a median of 58. \textit{Scene-Rule} has medium-length rule candidates, with a mean of 27.9 tokens and a median of 27. \textit{WordChecker} has very short candidates, with a mean of 2.7 tokens and a median of 3.

Within each benchmark, token lengths are tightly concentrated around their medians. Therefore, ordering candidates creates an approximately linear relationship between candidate index and absolute token position.

\section{Lexical Diversity: Measure of Textual Lexical Diversity (MTLD).} \label{app:mtld}

We compute lexical-diversity metrics over the candidate context only, excluding fixed prompt instructions and output-schema text. Tokens are whitespace-tokenized, lowercased, and stripped of punctuation before computing MATTR.

We also measure lexical diversity using the Measure of Textual Lexical Diversity (MTLD)~\citep{mccarthy2010mtld}. 
MTLD estimates the average length of a contiguous token span over which lexical variation remains above a fixed type-token ratio (TTR) threshold. 
Let $w_1,\dots,w_N$ be a token sequence and let $\tau$ denote the MTLD threshold, conventionally set to $\tau=0.72$~\citep{mccarthy2010mtld}. 
For a scan direction $d$, we move through the sequence and maintain the TTR of the current segment $w_s,\dots,w_j$:

\begin{equation}
\mathrm{TTR}(s,j)
=
\frac{|\{w_s,\dots,w_j\}|}{j-s+1}.
\end{equation}

Whenever $\mathrm{TTR}(s,j) \leq \tau$, the current segment is counted as one MTLD factor and a new segment begins at $w_{j+1}$. 
If the final segment does not reach the threshold, we add a fractional factor

\begin{equation}
\phi
=
\frac{1-\mathrm{TTR}_{\mathrm{final}}}{1-\tau},
\end{equation}

where $\mathrm{TTR}_{\mathrm{final}}$ is the TTR of the remaining segment. 
If $F_d$ is the resulting number of full plus fractional factors in direction $d$, then

\begin{equation}
\mathrm{MTLD}_d
=
\frac{N}{F_d}.
\end{equation}

Following standard practice, we compute MTLD in both the left-to-right and right-to-left directions and average the two scores:

\begin{equation}
\mathrm{MTLD}
=
\frac{1}{2}
\left(
\mathrm{MTLD}_{\rightarrow}
+
\mathrm{MTLD}_{\leftarrow}
\right).
\end{equation}

Larger MTLD values indicate that longer spans are needed before repetition lowers the TTR below $\tau$, and therefore correspond to greater lexical diversity.

\section{Ablation study: the impact of prompt truncation and post-target noise}
\label{app:post_target_ablation}

A common assumption in long-context processing is that fewer input tokens make retrieval easier, since the model has less to attend to. Removing tokens placed after the target should therefore improve performance. We test this assumption to see whether simple sorting can replace strict semantic filtering.

\begin{figure}[htpb]
  \centering
  \caption{\textbf{Effect of post-target noise on retrieval.} The charts compare success rates when the target is followed by a truncated prompt (filtering) versus a fully saturated $12k$-token context (sorting). Most frontier models handle the extra noise well, but a few degrade sharply under heavy distractor loads (e.g., Qwen models drop from 85\% to 50\% on \textit{WordChecker}).}

  \begin{subfigure}[b]{0.95\linewidth}
    \centering
    \includegraphics[width=\linewidth]{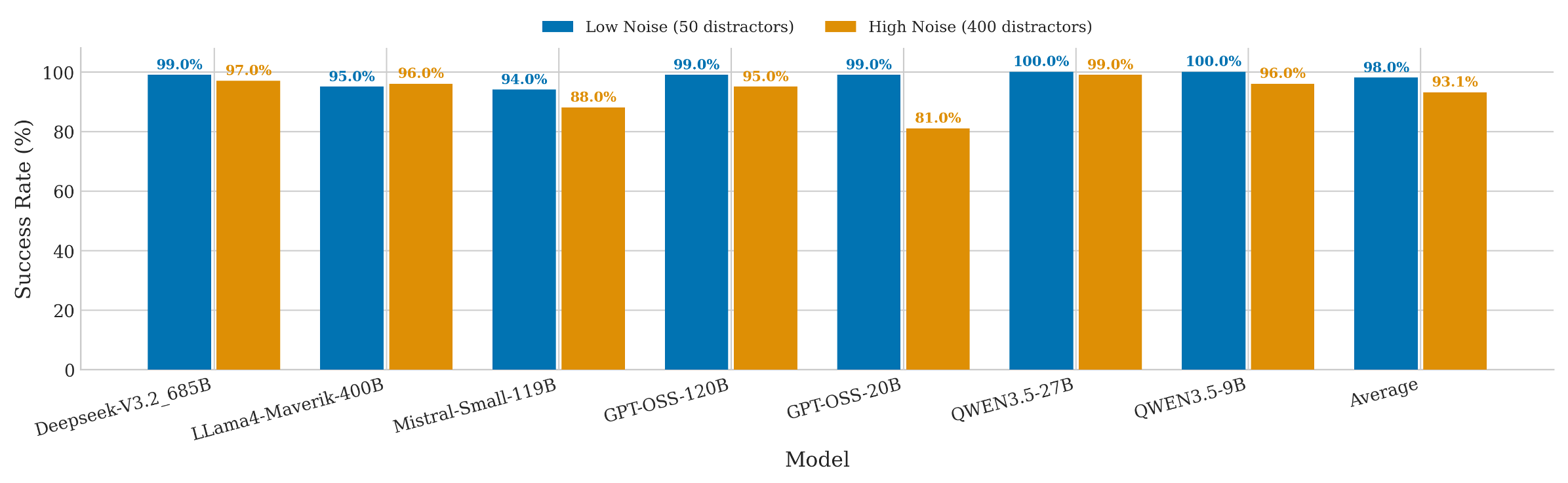}
    \caption*{(a) \textit{Scene-Rule}: truncated low-noise context (50 distractors) vs.\ saturated high-noise context (400 distractors).}
  \end{subfigure}

  \vspace{1em}

  \begin{subfigure}[b]{0.95\linewidth}
    \centering
    \includegraphics[width=\linewidth]{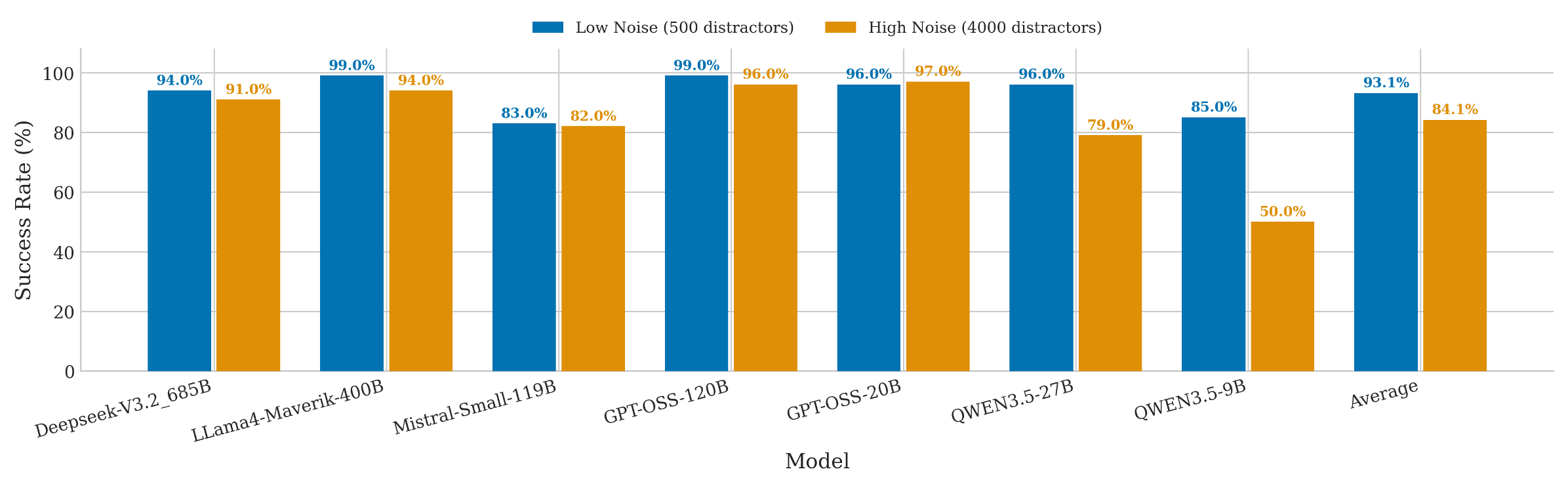}
    \caption*{(b) \textit{WordChecker}: truncated low-noise context (500 distractors) vs.\ saturated high-noise context (4000 distractors).}
  \end{subfigure}
  \label{fig:post_target_ablation}
\end{figure}

\paragraph{Experimental setup.}
As shown in Figure~\ref{fig:post_target_ablation}, we place the target near the beginning of the context and compare two conditions:
\begin{itemize}
    \item \textbf{Truncated context (semantic filtering):} The prompt ends shortly after the target, reducing the input to about $1.5k$ tokens (50 distractors in Scene-Rule, 500 in \textit{WordChecker}). This mimics a filter that drops irrelevant chunks.
    \item \textbf{Saturated context (sorting):} The target is placed early, but the prompt is padded to the full $12k$-token budget with distractors \textit{after} the target (400 in Scene-Rule, 4000 in \textit{WordChecker}). This mimics a system that ranks relevant chunks first but does not cut the context.
\end{itemize}
The setup checks whether a large block of tokens after the target still pulls attention away from information the model has already seen, and thus whether sorting alone is enough or filtering is strictly required.

\paragraph{Findings.}
Against expectations, most models cope well with downstream noise: success rates in saturated $12k$-token contexts are nearly identical to those in truncated, low-noise settings. In practice, this means simple sorting is enough for robust models. Their attention locks onto the target early and is not pulled away by later tokens, so strict semantic filtering is not needed.

There are, however, specific cases where the original intuition holds and truncation is still required. Qwen3.5 models degrade only on \textit{WordChecker} (e.g., Qwen3.5-9B drops from 85\% to 50\%): error logs show that high noise triggers an ``overthinking'' loop in which the model echoes the context until it runs out of tokens. GPT-OSS-20B degrades only on Scene-Rule (99\% to 81\%), likely because its smaller size struggles to keep attention anchored when surrounded by semantically dense pseudo-legal distractors.

\section{Detailed Error Analysis by Needle Position}
\label{app:errors}

As reported in our main findings, breaking down predictions on the \textit{WordChecker} benchmark under extreme lexical density reveals three distinct failure modes: \textbf{(i) Conflation}, where models mix candidate words with words from the query sentence; \textbf{(ii) Abstention}, where models refuse to answer; and \textbf{(iii) Loop Inversion}, where models enter a reasoning pattern that leads to output truncation.

\begin{figure*}[htpb]
    \centering
    \begin{subfigure}[b]{0.32\textwidth}
        \includegraphics[width=\textwidth]{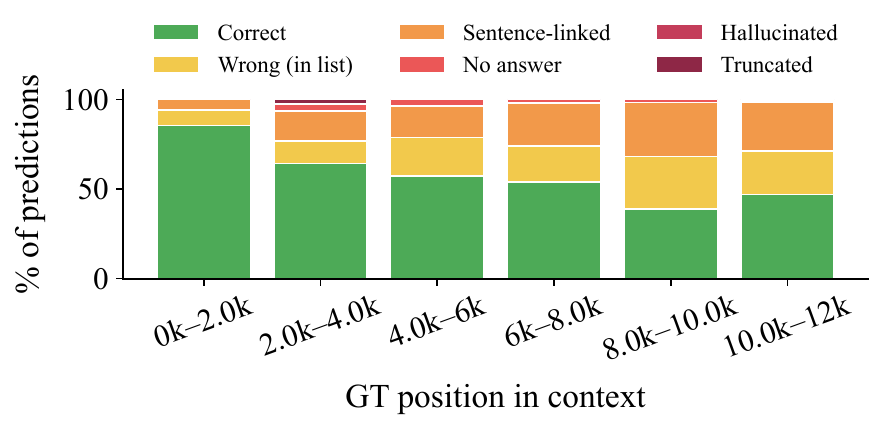}
        \caption{DeepSeek-V3.2}
    \end{subfigure}
    \hfill
    \begin{subfigure}[b]{0.32\textwidth}
        \includegraphics[width=\textwidth]{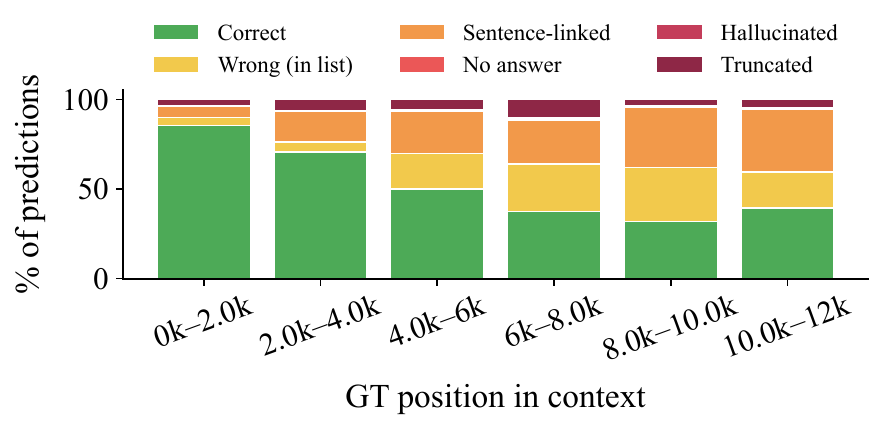}
        \caption{Llama-4 Maverick}
    \end{subfigure}
    \hfill
    \begin{subfigure}[b]{0.32\textwidth}
        \includegraphics[width=\textwidth]{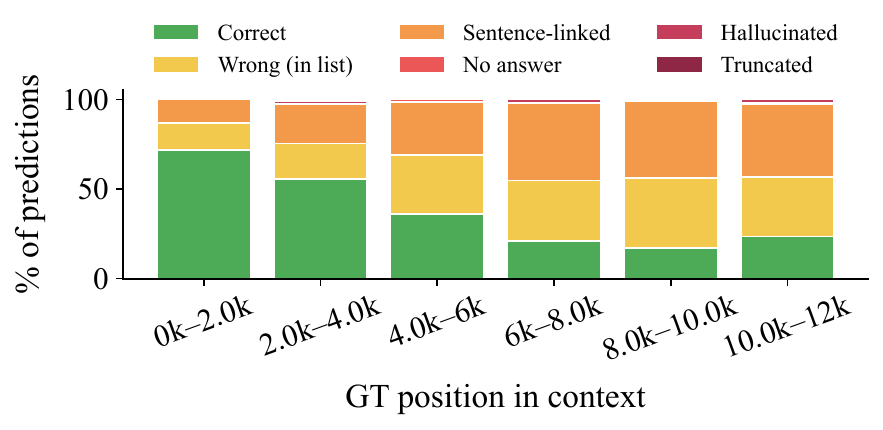}
        \caption{Mistral Small}
    \end{subfigure}

    \vspace{1.5em}

    \begin{subfigure}[b]{0.32\textwidth}
        \includegraphics[width=\textwidth]{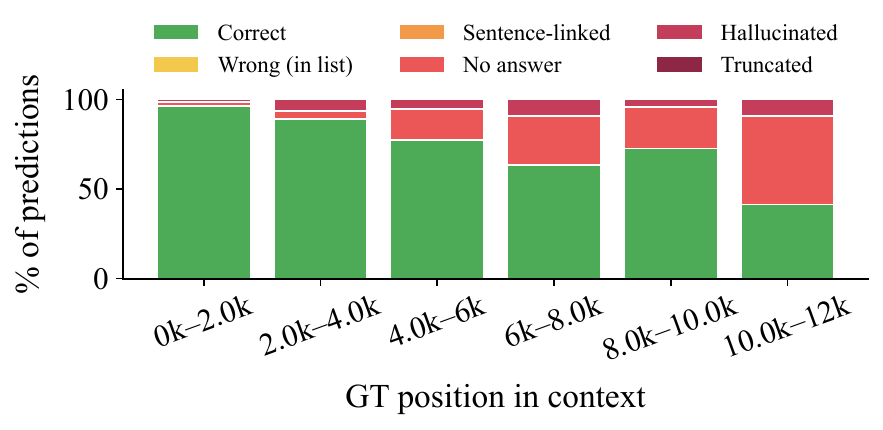}
        \caption{GPT-OSS-120B}
    \end{subfigure}
    \hfill
    \begin{subfigure}[b]{0.32\textwidth}
        \includegraphics[width=\textwidth]{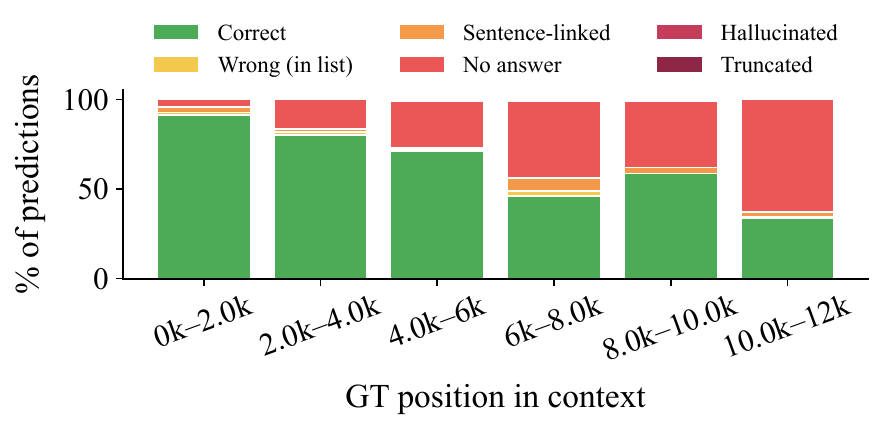}
        \caption{GPT-OSS-20B}
    \end{subfigure}
    \hfill
    \begin{subfigure}[b]{0.32\textwidth}
        \includegraphics[width=\textwidth]{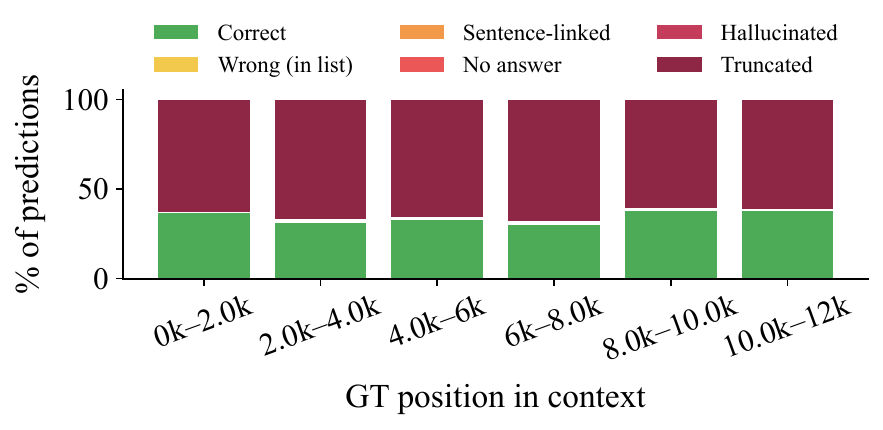}
        \caption{Qwen3.5-9B}
    \end{subfigure}

    \vspace{1.5em}

    \begin{subfigure}[b]{0.32\textwidth}
        \includegraphics[width=\textwidth]{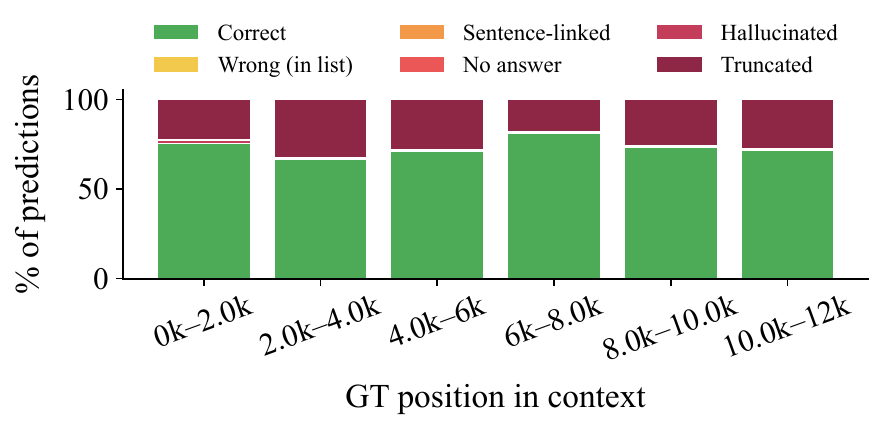}
        \caption{Qwen3.5-27B}
    \end{subfigure}
    \begin{subfigure}[b]{0.32\textwidth}
        \includegraphics[width=\textwidth]{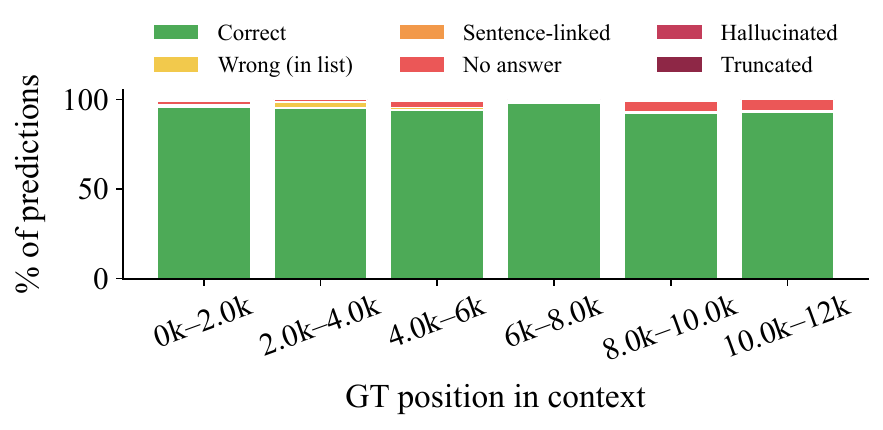}
        \caption{Qwen3.5-397B}
    \end{subfigure}
    \hspace{0.02\textwidth}

    \caption{\textbf{Prediction outcomes by ground-truth (GT) needle position on \textit{WordChecker}.} The stacked bars show that overall accuracy drops as the target is placed deeper in the context, but the specific failure mode (Conflation, Abstention, or Truncation) stays the same for each model across all positions.}
    \label{fig:detailed_error_bars}
\end{figure*}

To see how these failures behave with depth, we group predictions by the ground-truth (GT) position of the target needle within the saturated $12k$-token context. Figure~\ref{fig:detailed_error_bars} shows this position-wise breakdown for all evaluated models.

Two observations stand out:
\begin{enumerate}
    \item \textbf{Performance drops with depth.} As expected in needle-in-a-haystack settings, accuracy decreases the deeper the needle is placed. The share of correct predictions falls steadily across almost all models, raising the overall error rate.
    \item \textbf{The type of error does not change.} While the \textit{number} of errors grows with depth, the \textit{kind} of error stays the same for each model. Greater depth amplifies a model's existing failure mode rather than introducing new ones.
\end{enumerate}

For example, models in the \textbf{Conflation} regime (Mistral Small, DeepSeek-V3.2) lose their correct predictions at deeper bins entirely to ``Wrong (in list)'' and ``Sentence-linked'' false alarms; they do not start refusing to answer. Models in the \textbf{Abstention} regime (GPT-OSS variants) instead show a growing ``No answer'' block at deeper positions, without ever drifting into hallucinations or conflation. The Qwen family's \textbf{Loop Inversion} appears as a steady share of ``Truncated'' outputs that grows with depth as token exhaustion compounds.

Overall, the search strategies models use in hyper-dense contexts appear hardcoded: when pushed past their limits, they fail in predictable, architecture-specific ways.

\end{document}